\documentclass[a4paper,12pt]{article}
\usepackage[utf8]{inputenc}
\usepackage{lineno,hyperref}

\usepackage{textcomp}
\usepackage{array}
\usepackage{amsmath,amssymb} 
\usepackage{color}
\usepackage{diagbox, multirow}
\usepackage{makecell}
\usepackage{graphicx}
\usepackage{latexsym}
\usepackage{xcolor}
\usepackage{amsmath,amssymb}
\makeatletter
\newsavebox\myboxA
\newsavebox\myboxB
\newlength\mylenA

\newcommand*\xoverline[2][0.75]{%
	\sbox{\myboxA}{$\m@th#2$}%
	\setbox\myboxB\null
	\ht\myboxB=\ht\myboxA%
	\dp\myboxB=\dp\myboxA%
	\wd\myboxB=#1\wd\myboxA
	\sbox\myboxB{$\m@th\overline{\copy\myboxB}$}
	\setlength\mylenA{\the\wd\myboxA}
	\addtolength\mylenA{-\the\wd\myboxB}%
	\ifdim\wd\myboxB<\wd\myboxA%
	\rlap{\hskip 0.5\mylenA\usebox\myboxB}{\usebox\myboxA}%
	\else
	\hskip -0.5\mylenA\rlap{\usebox\myboxA}{\hskip 0.5\mylenA\usebox\myboxB}%
	\fi}
\makeatother
\usepackage{mathtools, geometry}
\usepackage{authblk}

\usepackage[caption=false,font=footnotesize]{subfig}

\begin{document}

\title{Low-Cost Self-Ensembles Based on Multi-Branch Transformation and Grouped Convolution}

\author[1]{Hojung Lee}
\author[1]{Jong-Seok Lee\thanks{Corresponding author: \texttt{jong-seok.lee@yonsei.ac.kr}}} 
\affil[1]{School of Integrated Technology, Yonsei University, Incheon 21983, South Korea}

\maketitle

\begin{abstract}
Recent advancements in low-cost ensemble learning have demonstrated improved efficiency for image classification. 
However, the existing low-cost ensemble methods show relatively lower accuracy compared to conventional ensemble learning. 
In this paper, we propose a new low-cost ensemble learning, which can simultaneously achieve high efficiency and classification performance. 
A CNN is transformed into a multi-branch structure without introduction of additional components, which maintains the computational complexity as that of the original single model and also enhances diversity among the branches' outputs via sufficient separation between different pathways of the branches. 
In addition, we propose a new strategy that applies grouped convolution in the branches with different numbers of groups in different branches, which boosts the diversity of the branches' outputs. 
For training, we employ knowledge distillation using the ensemble of the outputs as the teacher signal. 
The high diversity among the outputs enables to form a powerful teacher, enhancing the individual branch's classification performance and consequently the overall ensemble performance. 
Experimental results show that our method achieves state-of-the-art classification accuracy and higher uncertainty estimation performance compared to previous low-cost ensemble methods. 
The code is available at https://github.com/hjdw2/SEMBG.
\end{abstract}

\section{Introdution}
\label{sec:intro}

An ensemble of neural networks is an effective approach for enhancing the performance of deep neural networks for image classification \cite{ens1, ens2}.
One of the simplest ensemble methods is to combine the outputs of multiple models trained with different initializations, called Deep Ensembles \cite{Deepens}, which has been proven to be remarkably effective in enhancing classification performance compared to using a single model alone. 
As the model size or the amount of training data increases, however, this method becomes increasingly challenging since it demands substantial computational resources. 

In recent years, several ensemble learning methods with reduced computational overhead have emerged to mitigate this limitation as illustrated in Figs. 1a to 1c. 
They can be broadly categorized depending on the configuration of the ensemble members: using dropout or binary masks to generate multiple outputs from a single network \cite{MCdropout,Maskens} or employing cyclic learning rate schedules to generate multiple trained models \cite{Snapshot,FGE} (Fig. 1a); ensembling pruned sub-models \cite{PAT,Freeticket} (Fig. 1b); employing a single network structure that encapsulates multiple sub-networks \cite{Batchens,MIMO} or applying grouped convolution to a single network to obtain multiple outputs \cite{GEnet, PACKED} (Fig. 1c).
These methods are significantly more efficient than Deep Ensembles, but their classification performance usually falls short when compared to Deep Ensembles.

Enhancing ensemble performance relies on two critical factors: the quality of individual ensemble members and the diversity among them for complementarity \cite{divintro, Freeticket, PACKED}.
In this context, the approach in Fig. 1a can be appealing but is inefficient due to the necessity of multiple forward passes during inference.
The approach in Fig. 1b provides a more efficient solution, yet it may exhibit degradation in overall performance due to the reduced models.
The approach in Fig. 1c also introduces an efficient scheme, but the diversity among the ensemble members is limited because most of the network pathways are shared to generate each output.

\begin{figure}[t]
	\centering
	\includegraphics[height=3.0in]{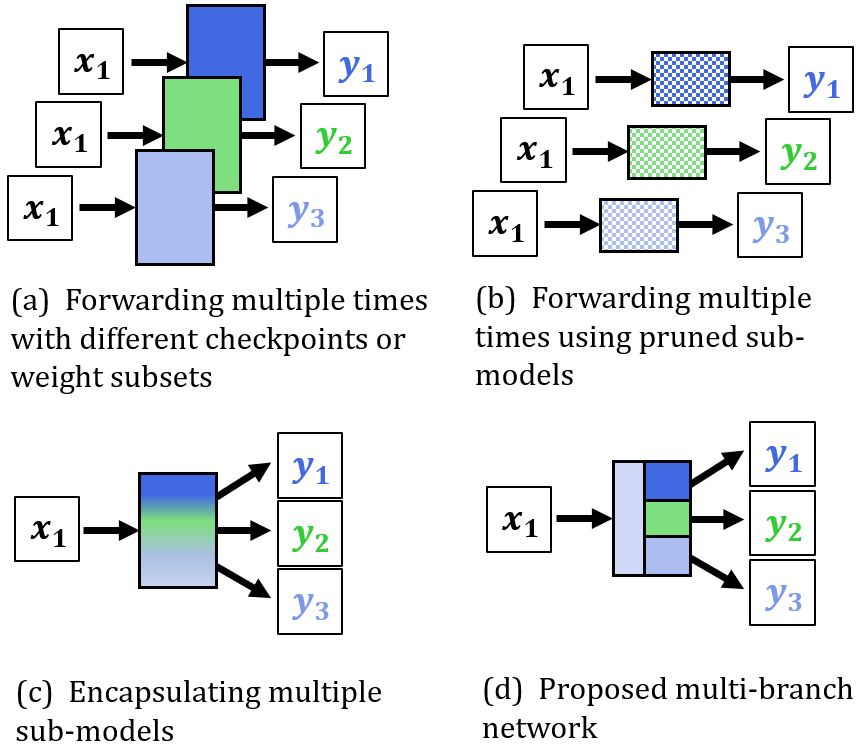} 
	\caption{Forward configurations for inference in the existing and proposed low-cost ensembles.}
	\label{Fig Intro}
\end{figure}

In this paper, we propose a new low-cost ensemble learning method using a multi-branch structure (Fig. 1d), which achieves high ensemble performance and diversity only with the computational cost of a single model.
To prevent the need for multiple forward passes during inference, a single model capable of producing multiple outputs concurrently is essential.
Furthermore, it is crucial to have sufficient separation between the pathways through which individual outputs are derived in order to ensure diversity among the outputs.
To fulfill these two objectives simultaneously, our method transforms a conventional CNN into a multi-branch structure by splitting the original network's layers into multiple sets to create multiple paths.
This ensures that the overall size of the transformed network remains similar to that of the original network, thereby making the computational burden for inference nearly unchanged from the original one.

In this multi-branch network, we propose a novel approach to employ grouped convolution \cite{Groupedconv} in the branches.
In \cite{GEnet, PACKED}, it is proposed to simply apply grouped convolution to a single conventional model and then consider the output of each group as an individual output. 
Grouped convolution inherently promotes feature diversity by enabling independent processing of each channel group.
To further capitalize on this characteristic, we propose a new strategy of assigning different numbers of groups for different branches. 
This deliberate variation in the number of groups allows each branch to generate outputs with enhanced diversity, effectively harnessing the potential of grouped convolution more explicitly than the approaches in \cite{GEnet, PACKED}.
Then, we create an ensemble output by aggregating the outputs of all branches and utilize it as a teacher signal for knowledge distillation-based training of the network.
Our group allocation strategy results in a superior ensemble teacher which is comprised of the outputs with high diversity, leading to enhanced performance of individual branches.
Consequently, these enhanced branches contribute to achieving improved ensemble performance compared to conventional methods.
We call our method Self-Ensembles using Multi-Branch and Grouped convolution (SEMBG).

To sum up, our method attains the three objectives as follows.
\begin{itemize}
	\item Efficiency: Our method transforms a CNN into a multi-branch structure without introducing additional components, so that the computational complexity remains almost the same to that of the original single model.
	\item Diversity: Each branch is well separated to ensure diversity between the branches. 
	In addition, our method assigns different numbers of groups for grouped convolution to each branch to further maximize the diversity.
	\item Classification performance: The enhanced diversity of the ensemble members helps compose a powerful teacher signal for knowledge distillation, which improves the performance of each branch and consequently the ensemble performance. 
\end{itemize}

We show that SEMBG achieves state-of-the-art classification accuracy and is also effective in uncertainty estimation when compared to recent ensemble methods on various datasets and network architectures.

The paper is organized as follows.
In Section 2, we delve into related work.
Section 3 provides a detailed description of the proposed SEMBG method.
Experimental results are presented in Section 4.
Lastly, Section 5 concludes the paper.

\section{Related Work}
\label{sec:related}
\subsection{Low-Cost Ensembles}

Following the success of Deep Ensembles \cite{Deepens}, low-cost methods have been studied to alleviate the high resource consumption.
TreeNet \cite{Treenet} introduces additional branches in the middle of a network to obtain multiple outputs.
Monte Carlo Dropout \cite{MCdropout} uses dropout \cite{Dropout} during inference to obtain multiple outputs.
BatchEnsemble \cite{Batchens} and Multi-Input Multi-Output Ensembles (MIMO) \cite{MIMO} encapsulate multiple sub-networks using shared weights with a rank-one matrix or using multi-input multi-output configuration, respectively.
Snapshot Ensembles (SSE) \cite{Snapshot} and Fast Geometric Ensembles (FGE) \cite{FGE} employ cyclic learning rate schedules, and HyperEnsembles \cite{Hyperensemble} employs different hyperparameter configurations to obtain multiple trained models.
FreeTickets Ensembles \cite{Freeticket} obtains sub-models using dynamic sparse training, and Prune and Tune Ensembles (PAT) \cite{PAT} obtains sub-models using anti-random pruning for ensembles.
Group Ensemble \cite{GEnet} and Packed-Ensembles \cite{PACKED} apply grouped convolution to a single network for obtaining multiple outputs.
However, while these methods demonstrate improved efficiency, their performance hardly reaches that of Deep Ensembles.


\subsection{Multi-Branch Architecture}

Neural networks with multi-branch have been studied for various purposes \cite{RelatedMB1,RelatedMB2}. 
Recently, they have been primarily used to improve classification performance utilizing the outputs from multiple branches \cite{ONE,okddip,MB3,MB2}.
The multi-branch structures in these works consume much more computational resources than the original single network because of the auxiliary branches. 
Therefore, direct performance comparison between these multi-branch networks and a single network would not be fair, as they involve different computational loads.
Furthermore, these methods utilize auxiliary branches to enhance the performance of the main branch, instead of the ensemble performance.
On the other hand, our goal is to implement a low-cost ensemble method that achieves significantly higher classification performance by ensembling, while maintaining the computational complexity similar to that of a single network.

\subsection{Grouped Convolution}

Grouped convolution divides the input and convolutional filters into groups and performs separate convolution, which can enhance both efficiency and feature diversity.
AlexNet \cite{Groupedconv} introduces grouped convolution to facilitate parallel computation of multiple independent convolution across multiple GPU devices.
After that, ResNeXt \cite{ResNext} demonstrates that using grouped convolution is beneficial for enhancing accuracy, and increasing the number of groups is more effective than using deeper or wider architectures under complexity constraints.
Group Ensemble (GENet) \cite{GEnet} and Packed-Ensembles \cite{PACKED} use grouped convolution for ensemble learning, which apply grouped convolution to a single network and treat the output of each group as an individual output.
However, this approach has limitations in terms of the diversity of the outputs since many features are shared during the process of obtaining the outputs. 
Therefore, we demonstrate that applying grouped convolution in a multi-branch network is a more effective approach for ensembling.

\begin{figure*}[t]
	\centering
	\includegraphics[height=2.0in]{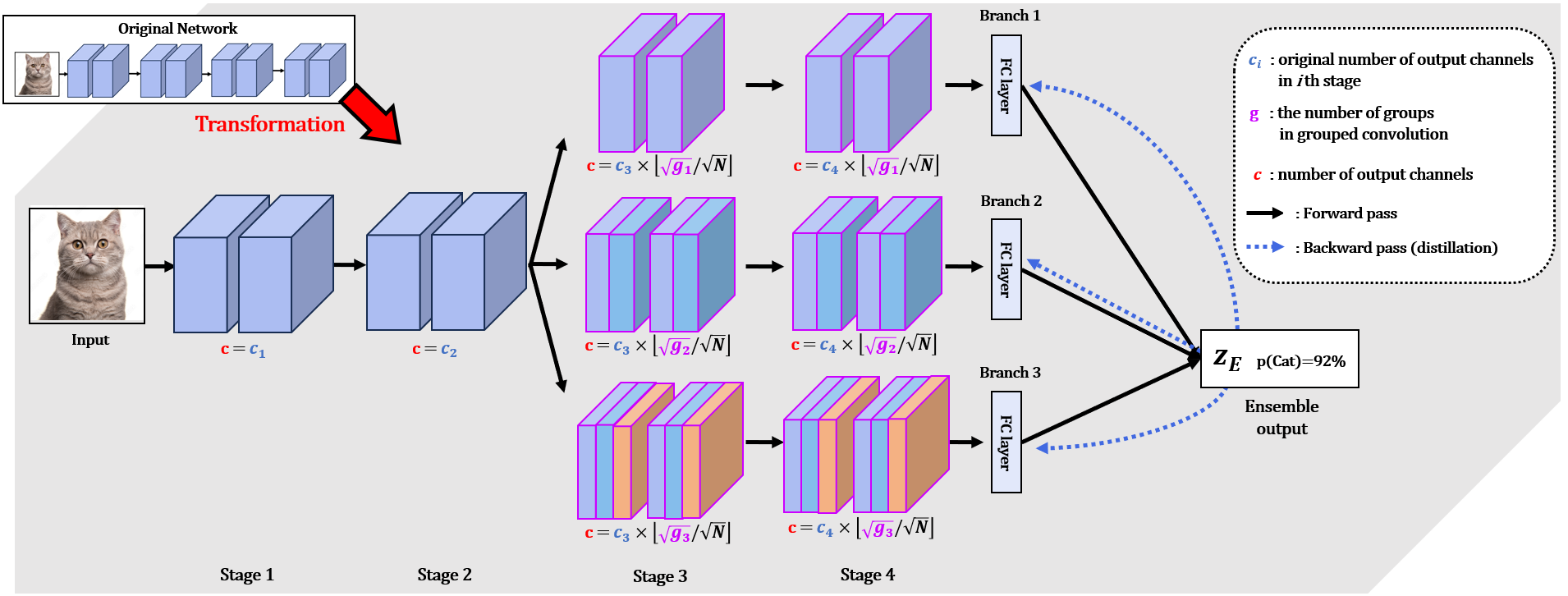} 
	\caption{Illustration of the proposed Self-Ensembles using Multi-Branch and Grouped convolution. 
		The boxes outlined with magenta color indicate the groups for grouped convolution. 
		In this illustration, $N$=3, $g_1$=1, $g_2$=2, and $g_3$=3.}
	\label{Fig SEMDG}
\end{figure*}

\section{Proposed Method}
\label{sec:method}

\subsection{Multi-Branch Network}
We begin by transforming an initialized CNN into a multi-branch structure.
In most existing CNN architectures, the same layer block is repeatedly used multiple times, which is commonly referred to as `stage'. 
A typical CNN comprises three (e.g. Wide-ResNet \cite{wrn}) or four (e.g., ResNet \cite{resnet}) stages.
Let $N$ be the number of ensemble members we want to have.
Then, we partition each layer within one stage into $N$ sets in the channel dimension to create $N$ branches as shown in Fig. \ref{Fig SEMDG}.
As in previous studies \cite{ONE,MB1,MB2}, the early stages are not partitioned but shared among the branches, which yields better performance.

Importantly, in order not to increase the computational burden, we keep the total size of the divided $N$ sets of layers to be the same to that of the original layers in the stage.
The number of weight parameters of a convolution layer is determined by $c_I \times c_o \times k^2$, where $c_I$ and $c_O$ are the numbers of input and output channels, respectively, and $k$ is the kernel size.
When a layer is divided into $N$ sets, we divide both $c_I$ and $c_O$ by $\sqrt{N}$ to match the size of the original layer:
\begin{equation}
c_I \times c_O \times k^2 \simeq N \times \left\lfloor \frac{c_I}{\sqrt{N}} \right\rfloor \times \left\lfloor \frac{c_O}{\sqrt{N}} \right\rfloor \times k^2.
\end{equation}
Thus, when the number of output channels of the $i$th layer in the original network is denoted as $c_O^i$, we set the number of output channels for each of the $N$ divided layers to $\frac{c_O^i}{\sqrt{N}}$.
Note that the number of input channels of the divided layers is automatically determined by the prior layer's output channels, i.e.,  $ \left\lfloor \frac{c_O^{i-1}}{\sqrt{N}} \right\rfloor = \left\lfloor \frac{c_I^i}{\sqrt{N}} \right\rfloor$.
This division process with adjusted output channel sizes ensures that the overall amount of computation remains nearly unchanged, making it a favorable solution for efficient ensemble learning without significant overhead.

\subsection{Grouped Convolution}
The multi-branch structure obtained above segregates different pathways well, which is beneficial to keep high diversity among the ensemble members. 
In order to further increase diversity, we propose employing grouped convolution in the branches.
Grouped convolution inherently promotes feature diversity by facilitating independent processing of different channel groups.
As a way of utilizing grouped convolution effectively, we introduce a strategy that assigns different numbers of groups for different branches.
When there are $N$ branches, we assign 1 to $N$ groups in each branch, respectively, which is straightforward but effective (see the experimental results).

When grouped convolution is applied in each branch, the number of weight parameters and the computational load reduce proportionally to the number of groups $g$ since the number of input channels for each filter is reduced by a factor of $g$.
This reduction may lead to degradation of the ensemble classification performance since the individual branch already has a decreased size compared to the size of the original path.
To compensate for this potential performance drop, we increase the number of input and output channels by a factor of the square root of the number of groups, i.e., $\sqrt{g}$. 
\begin{equation}
c_I \times c_O \times k^2 \simeq \left\lfloor \sqrt{g} \times \frac{c_I}{g} \right\rfloor \times \left\lfloor \sqrt{g} \times c_O \right\rfloor \times k^2.
\end{equation}
This adjustment ensures that the number of weight parameters remains similar between the cases when grouped convolution is utilized and when it is not. 
As a result, performance drop can be effectively prevented and, at the same time, a high level of feature diversity in the outputs of the branches can be obtained.

\subsection{Training}
The loss for training the whole network is composed of two components: cross-entropy loss and knowledge distillation loss. 
In the latter component, the ensemble of the outputs of all branches is used as the teacher signal \cite{MB1,MB2}. 
Due to the enhanced diversity of the branch outputs, knowledge distillation using the teacher signal combining these outputs is effective in improving the classification performance.

Given a training data $x$ and the corresponding one-hot label $y \in \mathcal{R}^M$ from $M$ classes, let the logit output of the $i$th branch be
\begin{equation}
z_i = f(x;\theta_i),
\end{equation}
where $f( \cdot; \theta_i )$ is the neural network implemented by the $i$th branch's pathway having weight parameters $\theta_i$.
The output probability of class $m$ is computed as
\begin{equation}
p_i^m = \frac{\exp(z_i^m)}{\sum_{j=1}^{M} \exp(z_i^j)},
\end{equation}
where $z_i^j$ is the $j$th element in $z_i$.
Then, the cross-entropy loss is computed as 
\begin{equation}
L_i^{CE} = \sum_{m=1}^{M} y^m \log p_i^m,
\end{equation}
where $y^m$ is the $m$th component of $y$.
Since we have $N$ outputs from the $N$ branches, i.e., $p_1^m$ to $p_N^m$, we have total $N$ cross-entropy losses. 
By summing them, the total cross-entropy loss is obtained:
\begin{equation}
L^{CE} = \sum_{i=1}^{N} L_i^{CE}.
\end{equation}

Next, for knowledge distillation, we use the Kullback-Leibler divergence as the loss.
Specifically, we use the average of the logit outputs from all branches as an ensemble teacher, which is written as
\begin{equation}
\label{eq1}
z_E = \frac{1}{N} \sum_{n=1}^{N} z_i.
\end{equation}
Then, we obtain the output probability of class $m$ using temperature $t$ as
\begin{equation}
\label{temp3}
p_E^m = \frac{\exp(z_E^m/t)}{\sum_{j=1}^{M} \exp(z_E^j/t)}.
\end{equation}
We also apply the temperature for obtaining students $p_i^m$.
Then, the distillation loss for the $i$th branch is given by
\begin{equation}
L_i^{KD} = t^2 \sum_{m=1}^{M} p_E^m \log \frac{p_E^m}{p_i^m}.
\end{equation}
As in the cross-entropy loss, the total distillation loss can be obtained by summing all the $N$ distillation losses:
\begin{equation}
L^{KD} = \sum_{i=1}^{N} L_i^{KD}.
\end{equation}

Finally, our loss is given by
\begin{equation}
\label{finaleq}
L = L^{CE} + \alpha L^{KD}.
\end{equation}
where $\alpha$ is a balancing coefficient.

For ensemble inference, the ensemble output is obtained by Eq. \ref{eq1}.

\section{Experiments}
\label{sec:exp}

In this section, we evaluate our proposed SEMBG.
First, we perform an ablation study on the configuration of the multi-branch structure.
Second, we evaluate the accuracy and the performance of uncertainty estimation of SEMBG compared to recent state-of-the-art low-cost ensemble learning methods.
For CIFAR100 \cite{cifar}, we use ResNet, Wide-ResNet, and DenseNet \cite{densenet}.
For CIFAR10, we use Wide-ResNet.
For ImageNet-1k \cite{imagenet}, we use ResNet50.
Lastly, we examine the diversity among ensemble members to investigate the underlying mechanism of the performance enhancement of our method.

We follow the default network structure and hyperparameter setting in \cite{PAT} for consistency in evaluation.
In addition, we use the pre-activation block for ResNet as proposed in \cite{preact}.
For Wide-ResNet, we do not use the dropout.
When a network is transformed into a multi-branch structure, the first stage (for three-stage networks) or the first two stages (for four-stage networks) are shared and the remaining stages are divided into branches.
By default, we utilize three branches (i.e., $N$=3), and they have different numbers of groups for grouped convolution: 1, 2, and 3, respectively.

We use the stochastic gradient descent (SGD) with a momentum of 0.9 and an initial learning rate of 0.1.
The temperature ($t$) in Eq. \ref{temp3} is set to 3.
For CIFAR100 and CIFAR10, the batch size is set to 128 and the maximum training epoch is set to 200.
The learning rate is decayed by a factor of 10 at half of the total number of epochs and then linearly decreases until 90\% of the total number of epochs, so that the final learning rate is 0.01 of the initial value.
The L2 regularization is used with a fixed constant of $5 \times 10^{-4}$. 
$\alpha$ in Eq. \ref{finaleq} is simply set to 1.
For ImageNet-1k, the batch size is set to 256 and the maximum training epoch is set to 150.
The learning rate decreases by an order of magnitude after 50, 100, and 130 epochs. 
The L2 regularization is used with a fixed constant of $1 \times 10^{-4}$. 
$\alpha$ in Eq. \ref{finaleq} is set to 0.1 because in this case individual branches show rather suboptimal classification performance and thus it is beneficial to reduce the influence of the knowledge distillation using the branches' outputs.

All experiments are performed using Pytorch with NVIDIA RTX 8000 GPUs.
We conduct all experiments three times with different random seeds and report the average results.

\subsection{Ablation Study}

We explore the classification performance depending on the branch configuration in our proposed SEMBG method. 
First, we analyze the performance with respect to the number of branches.
Next, we examine the effect of the assignment of the number of groups in each branch. 
Lastly, we experiment with different output aggregation methods used for obtaining an ensemble output.
In this ablation study, we utilize Wide-ResNet28-10 on CIFAR100.

\begin{figure*}[t]
	\centering
	\subfloat[Effect of the number of branches ($N$)]{%
		\includegraphics[width=0.48\linewidth]{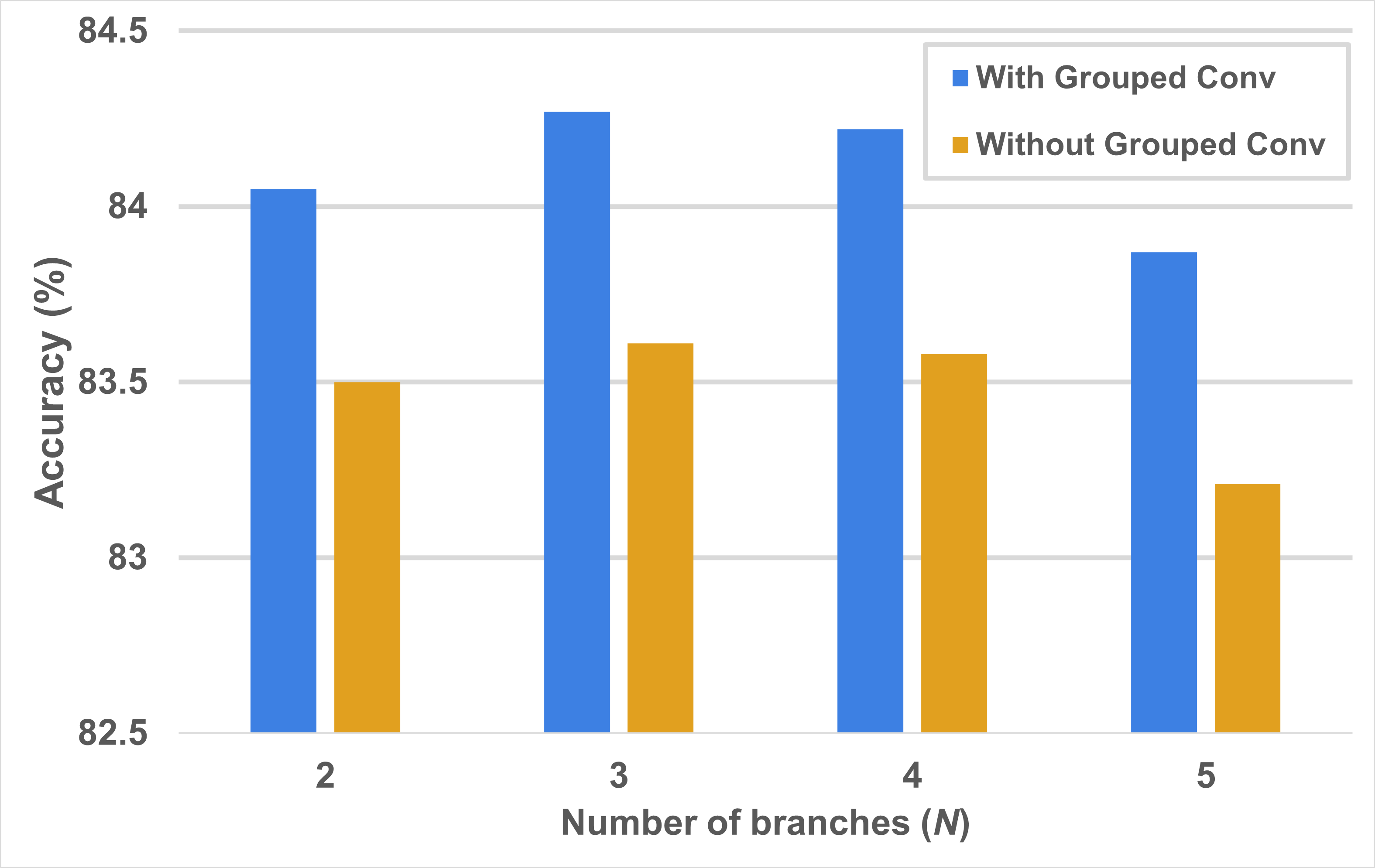}%
		\label{fig:ab1}
	}
	\hfill
	\subfloat[Effect of variation in the numbers of groups for grouped convolution]{%
		\includegraphics[width=0.48\linewidth]{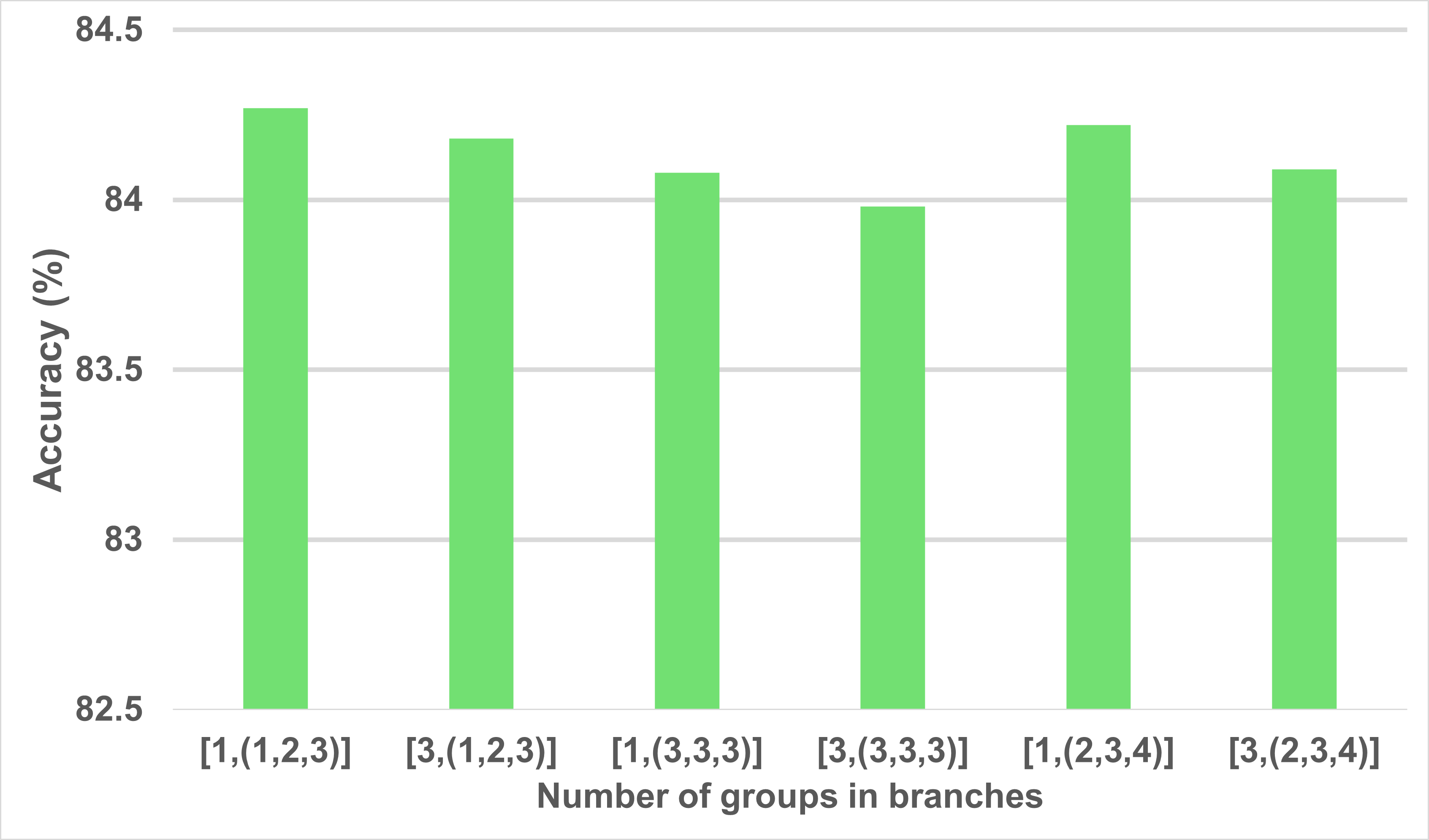}%
		\label{fig:ab2}
	}
	\vfill
	\subfloat[Effect of branch aggregation methods]{%
		\includegraphics[width=0.48\linewidth]{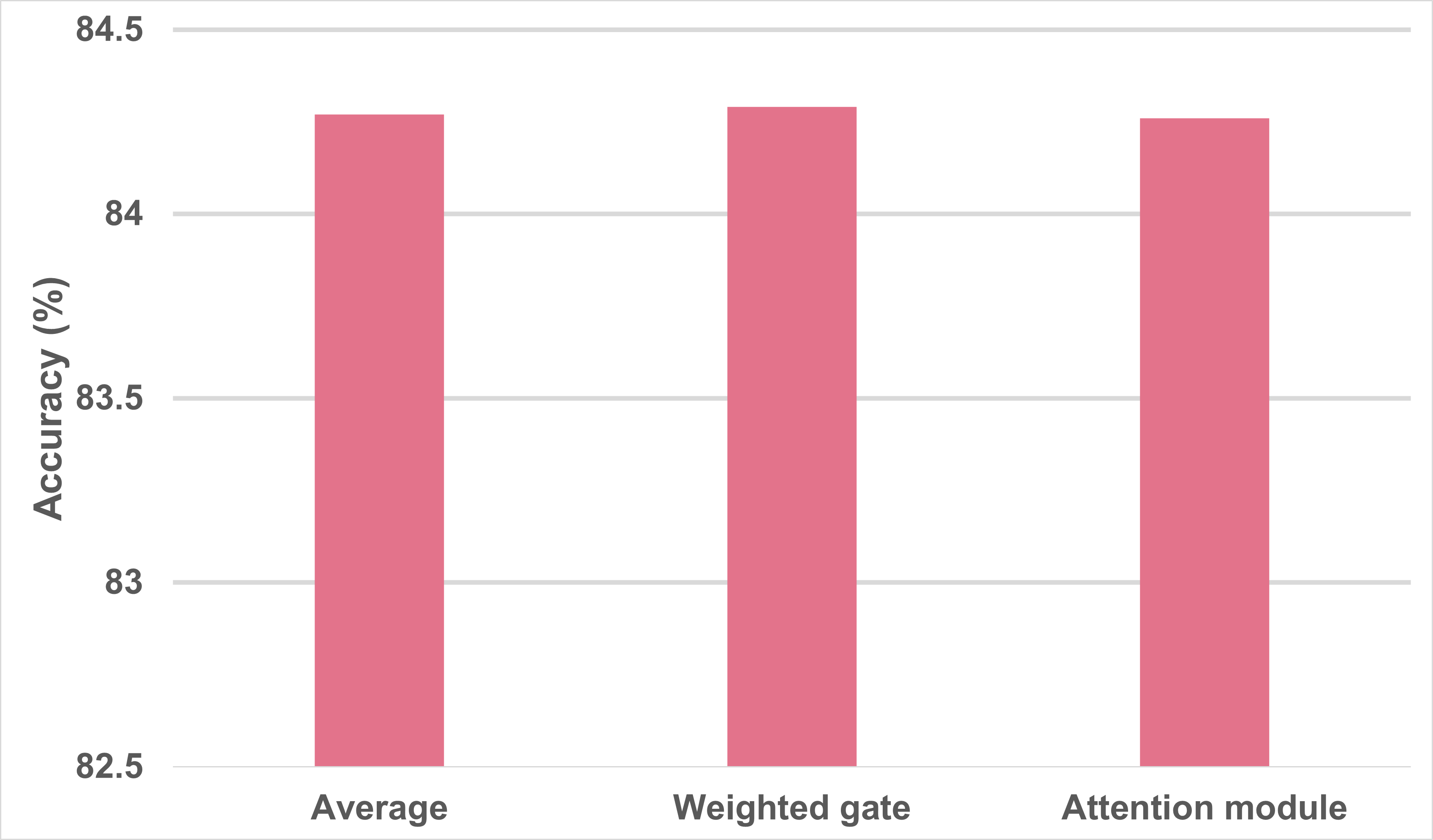}%
		\label{fig:ab3}
	}
	\caption{Effects of network configurations of SEMBG.}
	\label{Fig ab}
\end{figure*} 

\textbf{The number of branches}
The multi-branch network we employ offers the flexibility to adjust the number of ensemble members simply. 
During transformation, our method scales down the size of each branch in order to keep the computational burden unchanged. 
Accordingly, with an increase of the number of branches, the efficacy of ensemble learning may diminish due to the reduced learning capability of individual branches.
Thus, we perform an experiment to ascertain how many branches are appropriate for ensemble learning.

The results in Fig. \ref{fig:ab1} indicate that the best performance is achieved when three or four branches are used. 
Then, the accuracy starts to decrease as the number of branches increases further.
Therefore, we set the number of branches as three in all subsequent experiments.

Fig. \ref{fig:ab1} also includes the results when conventional convolution is applied instead of grouped convolution for reference, which exhibits a similar trend but overall with lower accuracy. 
This shows the effectiveness of grouped convolution in our method.

\textbf{Assignment of the numbers of groups in branches}
To enhance the diversity of each branch in our multi-branch structure, we assign different numbers of groups to each branch for grouped convolution. 
We perform an experiment to identify the most effective allocation strategy.
Here, we set the number of branches to three.
As shown in Fig. \ref{fig:ab2}, the notation [a,(b,c,d)] is used to denote the numbers of groups in the branches, where `a' represents the number of groups in the shared layers, and `b', 'c', and `d' denote the numbers of groups for the three branches, respectively. 
Note that when the number of groups is 1, grouped convolution reduces to conventional convolution.

The best performance is obtained with the proposed configuration, i.e., [1,(1,2,3)], where grouped convolution is not used in the shared layers, and the branches employ 1, 2, and 3 groups, respectively.
When we introduce grouped convolution in the shared layers using [3,(1,2,3)], the accuracy slightly decreases.
Next, we explore the cases where uniform group assignments are applied to all branches, i.e., [1,(3,3,3)] and [3,(3,3,3)]. 
These configurations also exhibit decreases in performance since the diversity among the branches is diminished.
Furthermore, when we increase the numbers of groups from the proposed configuration, i.e., [1,(2,3,4)] and [3,(2,3,4)], no meaningful improvement is observed.

\textbf{Output aggregation methods}
How to aggregate the outputs from individual branches for constructing an ensemble has been a primary subject in existing studies on the multi-branch structure \cite{ONE,okddip,MB2}. 
Hence, we conduct an experiment exploring the effect of different aggregation methods.
In particular, we compare the performance of the following three methods: the simple averaging of outputs, employing a linear layer to allocate weights to each branch's output \cite{ONE}, and using an attention module to assign a weight to the output of each branch \cite{okddip}.

Fig. \ref{fig:ab3} shows that no substantial variation in performance is observed across the methods. 
This is because although a different number of groups is assigned for each branch, their individual performance is similar to each other (81.7\%, 81.7\%, and 81.6\%, respectively).
Accordingly, assigning different weights to different branches does not bring a significant advantage in enhancing the ensemble performance.
Given these outcomes, we opt to utilize the simple averaging method instead of complicated aggregation approaches.

\begin{table*}[t] 
	\footnotesize
	\begin{center}
		{\caption{Performance comparison for CIFAR100 with three ensemble members ($N$=3). The best results are marked in bold and the second-best results with underlines.}
			\label{table1}}
		\begin{tabular}{lccccccccc}
			\hline
			Network & Method &Acc (\%) $\uparrow$&NLL $\downarrow$&ECE $\downarrow$& \makecell{FLOPs $\downarrow$ \\ (GMac)} & \makecell{Params $\downarrow$ \\ (M)} \\
			\hline
			\multirow{4}{*}{ResNet18} 
			& Single Model    & 79.1 & 0.853 & 0.067 & 0.73 & 11.22 \\
			& GENet           & 79.8 & 0.840 & 0.083 & 0.70 & 11.35 \\
			& Deep Ensembles  & \underline{81.6} & \underline{0.710} & \underline{0.051} & 2.19 & 33.66 \\
			& SEMBG           & \textbf{82.0} & \textbf{0.689} & \textbf{0.044} & 0.75 & 11.50 \\
			\hline
			\multirow{4}{*}{ResNet34} 
			& Single Model    & 80.0 & 0.840 & 0.074 & 1.33 & 21.34 \\
			& GENet           & 80.9 & 0.786 & 0.070 & 1.30 & 21.38 \\
			& Deep Ensembles  & \underline{82.5} & \underline{0.694} & \underline{0.055} & 3.99 & 64.02 \\
			& SEMBG           & \textbf{82.9} & \textbf{0.686} & \textbf{0.054} & 1.35 & 21.55 \\
			\hline
			\multirow{4}{*}{DenseNet169} 
			& Single Model    & 80.8 & 0.756 & 0.059 & 1.07 & 12.66 \\
			& GENet           & 81.9 & 0.712 & 0.061 & 1.05 & 12.78 \\
			& Deep Ensembles  & \textbf{82.7} & \underline{0.627} & \textbf{0.041} & 3.21 & 37.98 \\
			& SEMBG           & \underline{82.6} & \textbf{0.612} & \underline{0.050} & 1.12 & 12.84 \\
			\hline
			\multirow{4}{*}{WRN28-10} 
			& Single Model    & 81.6 & 0.748 & 0.049 & 5.96 & 36.55 \\
			& GENet           & 82.7 & 0.707 & 0.051 & 5.85 & 36.70 \\
			& Deep Ensembles   & \underline{83.5} & \underline{0.641} & \textbf{0.035} & 17.88 & 109.65 \\
			& SEMBG           & \textbf{84.3} & \textbf{0.622} & \underline{0.041} & 6.02 & 36.84 \\
			\hline
		\end{tabular}
	\end{center}
\end{table*}

\begin{table*}[t] 
	\footnotesize
	\begin{center}
		{\caption{Performance comparison for CIFAR100 with Wide-ResNet28-10. The best results are marked in bold and the second-best results with underlines. The results of the methods marked with * and ** are from \cite{PAT} and \cite{PACKED}, respectively.}
			\label{table3}}
		\begin{tabular}{lcccccccc}
			\hline
			\rule{0pt}{12pt}
			Method &Acc (\%) $\uparrow$&NLL $\downarrow$&ECE $\downarrow$& \makecell{FLOPs $\downarrow$ \\ (GMac)} & \makecell{Params $\downarrow$ \\ (M)}\\
			\hline
			Single Model& 81.6 & 0.748 & 0.049 & 5.96 & 36.55 \\
			TreeNet ($N$=3) & 82.5 & 0.681 & 0.043 & 15.08 & 92.47 \\
			GENet ($N$=3) & 82.7 & 0.707 & 0.051 & 5.85 & 36.70 \\
			SSE ($N$=5)*  & 82.1 & 0.661 & 0.040 & 29.80 & 182.75 \\
			PAT ($N$=6)*  & 82.7 & 0.634 & \textbf{0.013} & 17.88 & 109.65 \\
			MIMO ($N$=3)* & 82.0 & 0.690 & \underline{0.022} & 5.96 & 36.74 \\
			EDST ($N$=7)* & 82.6 & 0.653 & 0.036 & 6.97 & 42.76 \\
			DST ($N$=3)*  & 82.8 & \underline{0.633} & 0.026 & 6.02 & 36.92 \\
			BatchEnsemble ($N$=4)**& 82.3 & 0.835 & 0.130 & 23.81 & 36.65\\
			\makecell[l]{Packed-Ensembles \\ (CutMix) ($N$=4)**}  & \underline{83.9} & 0.678 & 0.089 & 5.95 & 36.62 \\	
			Deep Ensembles ($N$=3) & 83.5 & 0.641 & 0.035 & 17.88 & 109.65 \\
			SEMBG ($N$=3) & \textbf{84.3} & \textbf{0.622} & 0.041 & 6.02 & 36.84 \\
			\hline
			SEMBG (CutMix) ($N$=3) & \textbf{85.6} & \textbf{0.542} & \underline{0.019} & 6.02 & 36.84 \\		
			\hline
		\end{tabular}
	\end{center}
\end{table*}

\begin{table*}[t]
	\footnotesize
	\begin{center}
		{\caption{Performance comparison for CIFAR10 with Wide-ResNet28-10. The best results are marked in bold and the second-best results with underlines. The results of the methods marked with * and ** are from \cite{PAT} and \cite{PACKED}, respectively.}
			\label{table3_10}}
		\begin{tabular}{lcccccccc}
			\hline
			\rule{0pt}{12pt}
			Method &Acc (\%) $\uparrow$&NLL $\downarrow$&ECE $\downarrow$& \makecell{FLOPs $\downarrow$ \\ (GMac)} & \makecell{Params $\downarrow$ \\ (M)} \\
			\hline
			Single Model& 96.2 & 0.132 & 0.017 & 5.96 & 36.55 \\
			TreeNet ($N$=3) & 96.3 & 0.128 & 0.016 & 15.08 & 92.47 \\
			GENet ($N$=3) & 96.3 & 0.129 & 0.018 & 5.85 & 36.70 \\
			SSE ($N$=5)*  & 96.3 & 0.131 & 0.015 & 29.80 & 182.75 \\
			PAT ($N$=6)*  & \textbf{96.5} & \textbf{0.113} & \textbf{0.005} & 17.88 & 109.65 \\
			MIMO ($N$=3)* & \underline{96.4} & 0.123 & 0.010 & 5.96 & 36.74 \\
			EDST ($N$=7)* & \underline{96.4} & 0.127 & 0.012 & 6.97 & 42.76 \\
			DST ($N$=3)*  & \underline{96.4} & 0.124 & 0.011 & 6.02 & 36.92 \\
			BatchEnsemble ($N$=4)**& 95.6 & 0.206 & 0.027 & 23.81 & 36.59 \\
			Packed-Ensembles ($N$=4)** & 96.2 & 0.133 & \underline{0.009} & 4.06 & 19.35 \\
			Deep Ensembles ($N$=3) & \textbf{96.5} & 0.122 & 0.016 & 17.88 & 109.65 \\
			SEMBG ($N$=3) & \textbf{96.5} & \underline{0.121} & 0.015 & 6.02 & 36.84 \\		
			\hline
		\end{tabular}
	\end{center}
\end{table*}

\begin{table*}[t] 
	\footnotesize
	\begin{center}
		{\caption{Performance comparison for ImageNet-1k with ResNet50. The best results are marked in bold and the second-best results with underlines. The results of the methods marked with * are from \cite{PACKED}.}
			\label{table2}}
		\begin{tabular}{lccccccccc}
			\hline
			Method &Acc (\%) $\uparrow$&NLL $\downarrow$&ECE $\downarrow$& \makecell{FLOPs $\downarrow$ \\ (GMac)} & \makecell{Params $\downarrow$ \\ (M)} \\
			\hline
			Single Model                 & 77.7 & 0.951 & \underline{0.079} & 25.6 & 4.12 \\
			BatchEnsemble ($N$=4)*      & 75.9 & -     & \textbf{0.035} & 25.7 & 16.36 \\
			MIMO ($N$=4)*                & 77.6 & -     & 0.147 & 31.7 & 4.45 \\
			Packed-Ensembles ($N$=4)*   & \underline{77.9} & -     & 0.180 & 59.1 & 9.29 \\
			Deep Ensembles ($N$=2)       & \textbf{78.9} & \textbf{0.884} & 0.178 & 51.2 & 8.24 \\
			SEMBG ($N$=2)                & \underline{77.9} & \underline{0.947} & 0.121 & 39.9 & 5.39 \\
			\hline
		\end{tabular}
	\end{center}
\end{table*}

\subsection{Performance Comparison}

We first compare our SEMBG with Deep Ensembles on CIFAR100 for several kinds of networks, including ResNet, Wide-ResNet, and DenseNet.
We also compare ours with GENet \cite{GEnet}, which uses grouped convolution.
The primary performance measure is the accuracy (Acc) in \%, but we also use the Negative Log-Likelihood (NLL) and the Expected Calibration Error (ECE) \cite{ece} which evaluates the performance of uncertainty estimation.
In order to demonstrate the computational burden and efficiency, we also report the total number of floating point operations (FLOPs) and the number of parameters required for obtaining an ensemble output in the inference phase.

The results are shown in Table \ref{table1}.
In most cases, our SEMBG achieves better results than Deep Ensembles.
For both ResNet18 and ResNet34, SEMBG achieves the best results in all three performance metrics.
Our method exhibits a clear accuracy improvement (+0.4\%) over Deep Ensembles for both ResNet cases.
Furthermore, when compared to the single model, our method incurs only a minimal increase (+0.02GMac) in FLOPs, yet achieves a remarkable increase in accuracy (+2.9\%).
The most notable accuracy increase is observed in the case of Wide-ResNet28-10, surpassing Deep Ensembles by +0.8\%. 
Despite this significant improvement, the computational cost increases only marginally (+0.06GMac) compared to the single model.
In addition, SEMBG achieves great performance across all networks in terms of NLL and ECE.
The NLL and ECE of SEGMB are nearly on par with Deep Ensembles (with slight improvements overall).
However, the distinct advantage of SEGMB lies in its computational cost, which is approximately three times lower than Deep Ensembles.
When compared to GENet, SEMBG achieves consistently superior performance across all cases.

Next, we compare the performance of SEMBG with other state-of-the-art low-cost ensemble learning methods, including TreeNet \cite{Treenet}, SSE \cite{Snapshot}, PAT \cite{PAT}, MIMO \cite{MIMO}, Dynamic Sparse Training (DST), Efficient Dynamic Sparse Training (EDST) \cite{Freeticket}, BatchEnsemble \cite{Batchens}, and Packed-Ensembles \cite{PACKED}.
Table \ref{table3} summarizes the results for CIFAR100 with Wide-ResNet28-10.

As presented in Table \ref{table3}, our SEMBG method achieves the highest accuracy and NLL scores among all the evaluated methods. 
As mentioned in the introduction, while most existing methods can reduce computational cost compared to Deep Ensembles, they tend to exhibit lower accuracy. 
The only method surpassing Deep Ensembles in accuracy is Packed-Ensembles. 
However, we note that this comparison is not entirely fair since Packed-Ensembles employs data augmentation methods such as mixup \cite{mixup} and CutMix \cite{cutmix} to achieve these results while the other methods do not. 
In contrast, our SEMBG method achieves superior accuracy to Deep Ensembles without relying on any data augmentations. 
When the data augmentation methods, such as CutMix, are applied to SEMBG, we observe further performance boosts as shown in the last row of Table \ref{table3}.
Additionally, the computational cost difference between SEMBG and Packed-Ensembles is negligible (a relative difference of 1.1\%). 
Therefore, we assert that our method represents the best low-cost ensemble learning approach in the state-of-the-art.
In terms of ECE, our method also shows sufficiently good performance, which is a sharp contrast to Packed-Ensembles showing significantly degraded performance.

The performance comparison results for CIFAR10 with Wide-ResNet28-10 are presented in Table \ref{table3_10}. 
Compared to CIFAR100, for CIFAR10, the performance gap between the methods is relatively small but our SEMBG method achieves the highest accuracy and also attains sufficiently good performance in terms of NLL and ECE.
When compared with the methods that also attain the highest accuracy (i.e., PAT and Deep Ensembles), SEMBG has a clear advantage in computational efficiency.

For ImageNet-1k, while the performance of SEMBG might not stand as the absolute best, SEMBG shows satisfactory performance compared to other methods as shown in Table \ref{table2}.
Notably, SEMBG achieves the same accuracy with Packed-Ensemble but requires reduced computational complexity.

\subsection{Diversity Analysis}
In our pursuit of analyzing the performance enhancement achieved by SEMBG, we investigate the diversity exhibited among the branches. 
To this end, we assess the pairwise diversity using two metrics: prediction disagreement and cosine similarity.
The former is the ratio of test samples that two branches classify differently \cite{divintro}. 
The latter measures the cosine similarity of the softmax outputs from two branches \cite{divcos}.
We also report the average accuracy of ensemble members.

The results are presented in Table \ref{table7}.
SEMBG without knowledge distillation ($L^{KD}$) shows high diversity levels that are even higher than those of Deep Ensembles, which is remarkable in that Deep Ensembles produces completely independent multiple networks while SEMBG generates only one network yielding multiple outputs. 
Such high diversity enables to form a powerful teacher for distillation, which leads to high performance of individual branches in SEMBG (with distillation). 
Through distillation, the similarity among branches inevitably increases during training.

\begin{figure*}[t]
	\centering
	\subfloat[Prediction disagreement]{%
		\includegraphics[width=0.44\linewidth]{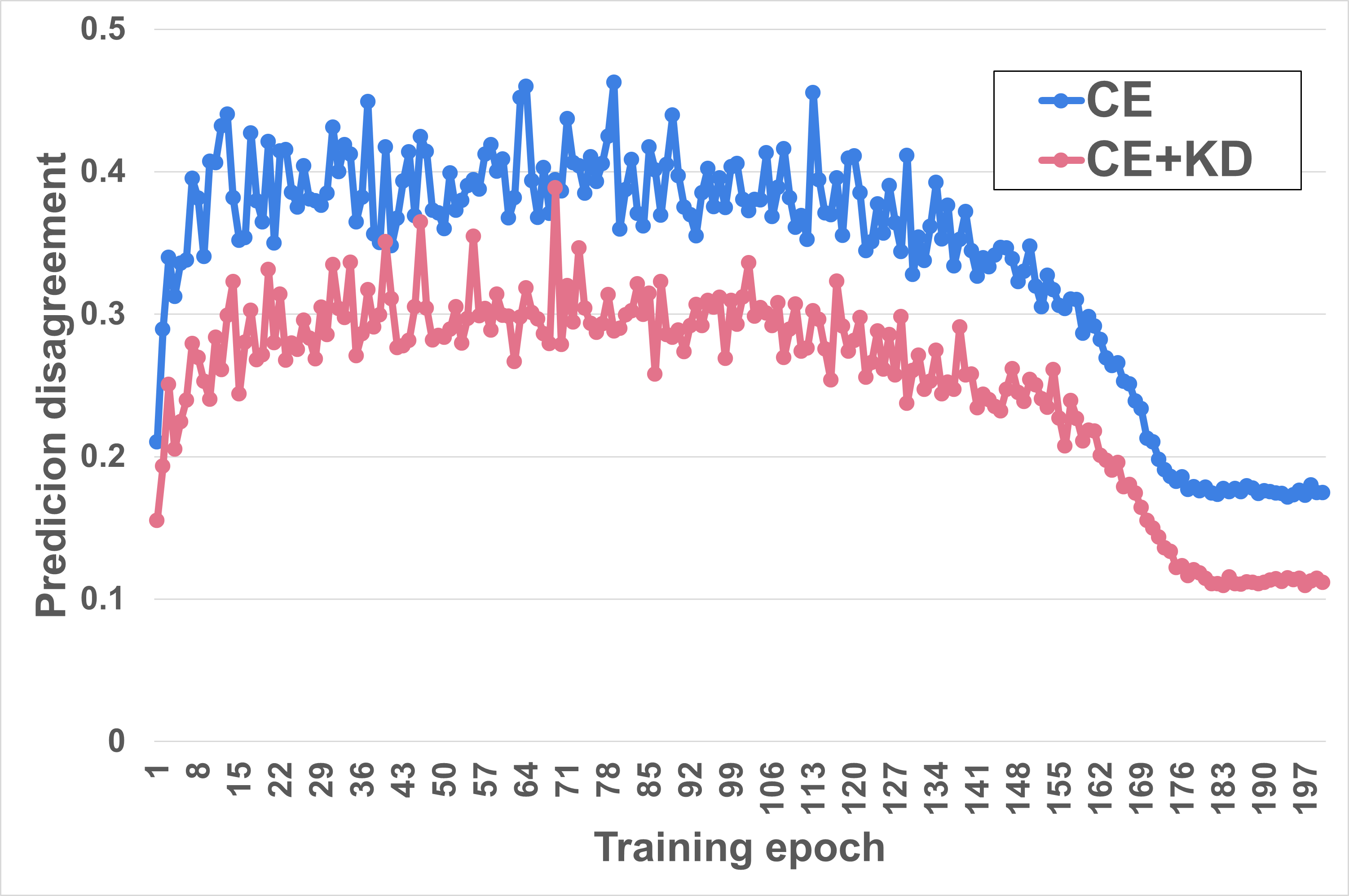}%
		\label{fig:div1}
	}
	\hfill
	\subfloat[Cosine similiarity]{%
		\includegraphics[width=0.44\linewidth]{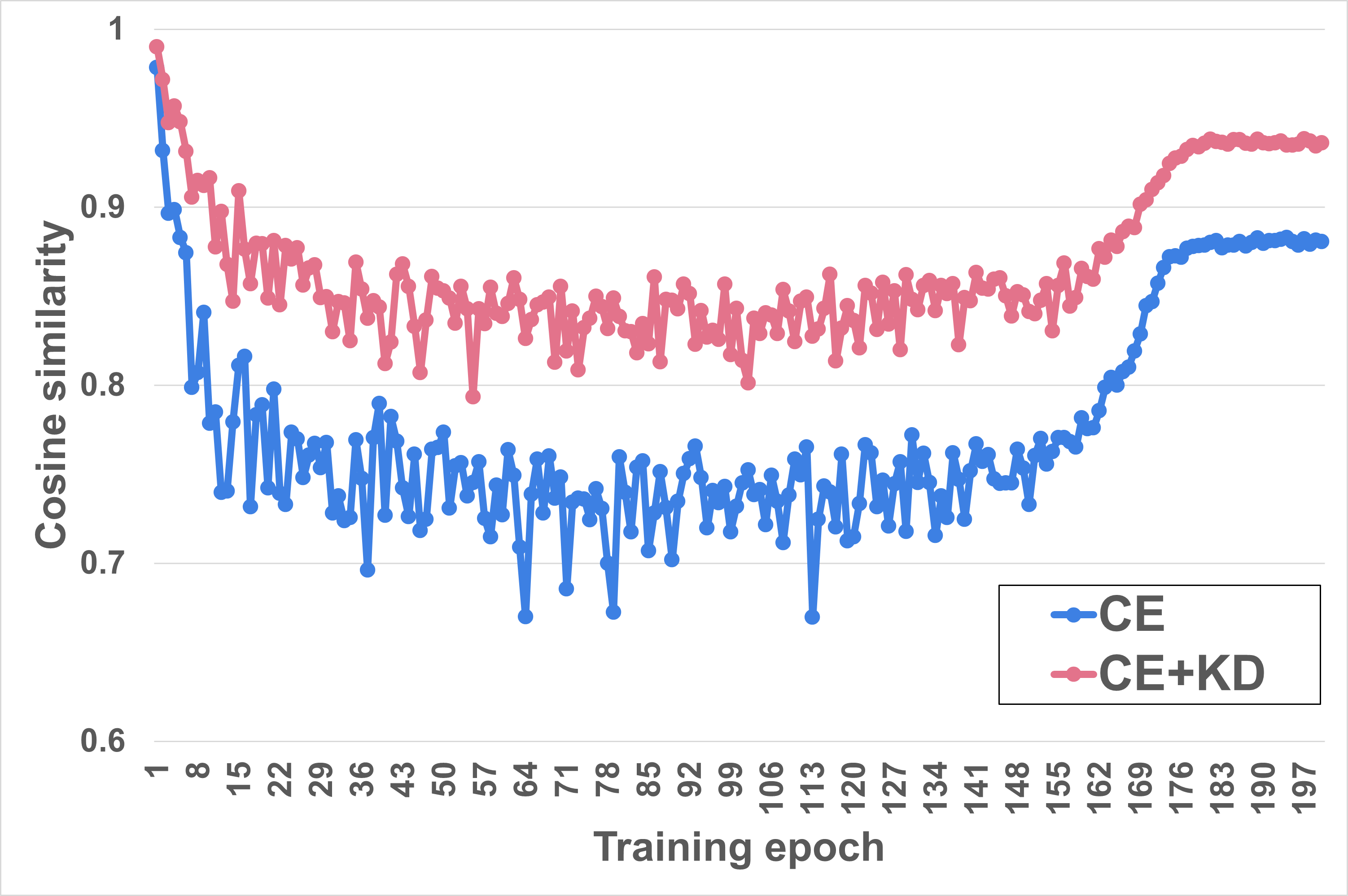}%
		\label{fig:div2}
	}
	\vfill
	\subfloat[Ensemble accuracy]{%
		\includegraphics[width=0.44\linewidth]{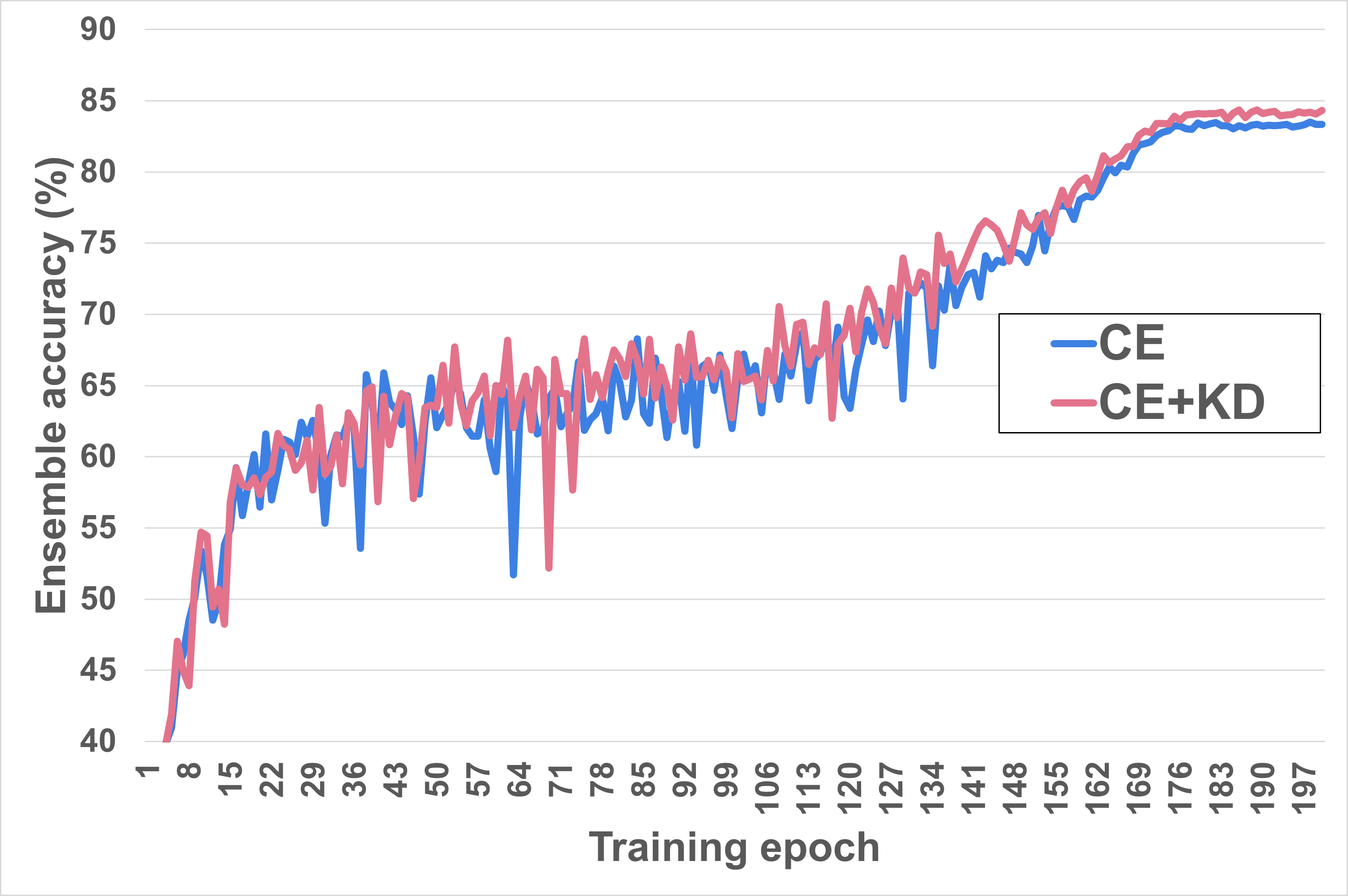}%
		\label{fig:div3}
	}
	\caption{Evolution of diversity among the branches in SEMBG during training for CIFAR100 with Wide-ResNet28-10.}
	\label{Fig div}
\end{figure*}

To further investigate the diversity of SEMBG, the diversity change during training is shown in Fig. \ref{Fig div}.
As training advances, diversity quickly increases, maintains a certain level for most of the training period, and eventually starts to decrease at the end of training.
The established diversity during training leads to enhancement in the performance of each branch through the ensemble teacher.
Without distillation (the blue curves in the figure), diversity is higher compared to the case with distillation (the red curves) but the ensemble performance is relatively low due to the lower performance of the individual branches.


\begin{table}
	\footnotesize
	\begin{center}
		{\caption{Prediction disagreement (PD) and cosine similarity (CS) between ensemble members, and average accuracy of ensemble members for CIFAR100 with Wide-ResNet28-10. A high PD or a low CS indicates a high level of diversity.}
			\label{table7}}
		\begin{tabular}{lcccccccc}
			\hline
			\rule{0pt}{12pt}
			$N$ & Method & PD & CS & Acc (\%) \\
			\hline
			\multirow{4}{*}{2} &
			GENet            & 0.1313 & 0.9179 & 80.9 \\
			& Deep Ensembles & 0.1514 & 0.9026 & 81.8 \\
			& SEMBG (w/o $L^{KD}$)& 0.1746 & 0.8807 & 81.7 \\
			& SEMBG          & 0.1125 & 0.9376 & 82.4 \\
			\hline
			\multirow{4}{*}{3} &
			GENet            & 0.1335 & 0.9198 & 81.1 \\
			& Deep Ensembles & 0.1501 & 0.9007 & 81.7 \\
			& SEMBG (w/o $L^{KD}$)& 0.1718 & 0.8846 & 81.8 \\
			& SEMBG          & 0.1131 & 0.9351 & 82.6 \\
			\hline
		\end{tabular}
	\end{center}
\end{table} 

\section{Conclusion}
\label{sec:conclusion}
We introduced a new low-cost ensemble learning approach using a multi-branch structure and grouped convolution.
An efficient multi-branch structure was obtained by transformation of a CNN.
Through the application of grouped convolution with the effective group assignment strategy, our network acquired outputs with high diversity from the branches, achieving high ensemble performance.
Our method showed state-of-the-art results that surpassed other recent methods, with only minimal additional computational overhead.
Further experiments for larger models or other tasks will be addressed in future work. 

\bibliographystyle{plain}
\bibliography{egbib}

\end{document}